\documentclass[conference]{IEEEtran}

\usepackage[utf8]{inputenc} 
\usepackage[english]{babel}
\usepackage[T1]{fontenc}  

\usepackage{epsfig}           
\usepackage{subfigure}
\usepackage{graphicx}
\usepackage{setspace}
\usepackage{amsmath}
\usepackage{amsfonts}
\usepackage{amssymb}
\usepackage{calligra}
\usepackage{eucal}
\usepackage[linesnumbered]{algorithm2e}

\usepackage{lscape}

\IEEEoverridecommandlockouts

\begin {document}

\title{Segmentation-Free Approaches for Handwritten Numeral String Recognition}

\author{\IEEEauthorblockN{Andre G. Hochuli, Luiz  S. Oliveira}
	\IEEEauthorblockA{Federal University of Parana\\
		Department of Informatics (DInf)\\
		Curitiba, PR - Brazil\\
		Email:  \{aghochuli, lesoliveira\}@inf.ufpr.br}
	\and
	\IEEEauthorblockN{Alceu  de Souza Britto Jr.}
	\IEEEauthorblockA{Pontifical Catholic University of
		Parana\\
		PPGIA \\
		Curitiba, PR - Brazil\\
		Email: alceu@ppgia.pucpr.br}
	\and
	\IEEEauthorblockN{Robert Sabourin}
	\IEEEauthorblockA{Ecole de Technologie Superieure\\
		Montreal, QC - Canada\\
		Email: robert.sabourin@etsmtl.ca}

	}

\maketitle

\begin{abstract}
This paper presents segmentation-free strategies for the recognition of handwritten numeral strings of unknown length. A synthetic dataset of touching numeral strings of sizes 2-, 3- and 4-digits was created to train end-to-end solutions based on Convolutional Neural Networks. A robust experimental protocol is used to show that the proposed segmentation-free methods may reach the state-of-the-art performance without suffering the heavy burden of over-segmentation based methods. In addition, they confirmed the importance of introducing contextual information in the design of end-to-end solutions, such as the proposed length classifier when recognizing numeral strings.    
\end{abstract}

\IEEEpeerreviewmaketitle

\section{Introduction}

\IEEEPARstart{T}he challenge of recognizing numeral strings of unknown length which are not neatly written is still an open problem in the field of document analysis and recognition. The difficulties contributing to the unsatisfactory performance of many methods available in the literature are usually related to the presence of broken, overlapping and touching digits in the string. In such a case, a straightforward solution to segment the string into components representing single digits usually becomes unfeasible. 

\begin{figure}[htbp]
	\begin{center}
		\mbox{
			\subfigure[]{\scalebox{0.45}{\epsffile{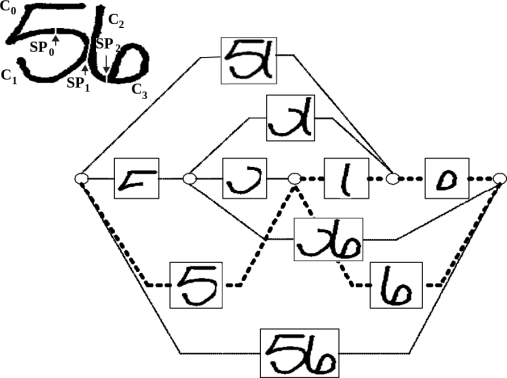}}} \quad
			\subfigure[]{\scalebox{0.20}{\epsffile{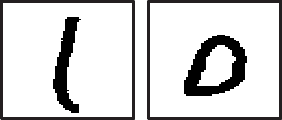}}}
		}
		\caption{(a) Segmentation paths for the string ``56'' and (b) Images that can be easily confused with digits ``1'' and ``0'' (extracted from \cite{Vellasques2008}). }
		\label{seggraph:fig}
	\end{center}
	\vspace{-7mm}
\end{figure}

One may find in the literature a variety of segmentation algorithms that explore different background and foreground features to provide potential segmentation cuts for a given unknown length numeral string. An interesting comparison of different approaches is presented in \cite{Ribas2013}, in which it is possible to observe that an alternative to reduce the heuristics necessary to provide the correct string segmentation cuts is the over-segmentation based algorithms. They segment the string, as many as necessary, into components that may represent digits or part of them. After obtaining the recognition result of each component, or their combination, the algorithms in this approach compute the optimal integrated result. The over-segmentation process for touching numeral `56' is depicted on Figure \ref{seggraph:fig}. In fact, the rational behind this approach is to maximize the chances of generating the correct segmentation cuts, however paying the high price of increasing significantly the computational cost of the segmentation/recognition process. 

The alternative methods resort to segmentation-free based methods in which the string is recognized without the need of its a priori segmentation into isolated digits. Such an approach has recovered the attention of the research community in the last years with the recent advances in machine learning motivated by the deep learning techniques. While the over-segmentation based methods demand some specific strategy to generate segmentation cuts, a robust isolated digit recognizer and a strategy for searching the best path among the generated segmentation hypothesis, the segmentation-free demands a significant amount of training data. 

One of the first attempts to apply Convolutional Neural Networks (CNN's) to recognize large input fields with unsegmented characters was done by Matan et al. \cite{Matan92}. For a given vector sequence, their SDNN (Spatial Displacement Neural Network) provides a series of output vectors that are post-processed in order to find out the best possible label sequence. The authors observed 66\% of correct classification on 3000 images of ZIP Codes. Even being an important contribution, the SDNN did not provide better results than the segmentation methods, as observed by LeCun et al. \cite{LeCun98}.

Another segmentation-free strategy was presented by Choi and Oh \cite{Choi99}. The authors trained a modular neural network composed of 100 separate subnetworks. They reported a 95.3\% recognition rate on 1374 digit pairs extracted from the NIST dataset. In a similar strategy, Ciresan \cite{Ciresan2008} trained a 100-class CNN using 200,000 images, and reported a 94.65\% recognition rate. In addition, he described experiments on 3-digit strings using two CNNs, one for isolated digits and the other for touching pairs. Spite of the fact that this work does not consider three overlapping digits, a 93.4\% recognition performance was reported on 1,476 3-digit strings from the NIST dataset. 

The aforementioned segmentation-free strategies rests on the assumption that most touching occurs between two adjacent digits. In the same direction, a quite recent approach was proposed by Hochuli et al. \cite{Hochuli2018}, which consists of a segmentation-free method based on dynamic selection of classifiers \cite{BRITTO2014}\cite{CRUZ2018}. The first one, named $\mathcal{L}$, is applied to estimate the number of components in the string, while other three are responsible to discriminate 10 [$0 \ldots 9$], 100 [$00 \ldots 99$], and 1000 [$000 \ldots 999$] classes. Their approach achieved state-of-art levels, surpassing segmentation based approaches for touching components.

Although Hochuli et al. \cite{Hochuli2018} brings up a new perspective to the problem, towards an end-to-end solution two important questions are still open: (a) Could a single classifier, capable to discriminate those 1110 classes, surpass the proposed dynamic selection strategy based on four classifiers? and, in this case, (b) May the string length classifier remain useful in a single based classifier method? 

To answer those questions we have implemented end-to-end solutions composed of a 1110 digit classifier combined with the string length classifier. In order to assess those approaches, we used a robust experimental protocol on a Touching Pairs (TP) dataset of 79,464 touching digits, as well as on Synthetic Dataset composed of 570,461 samples of isolated digits and touching strings of 2- and 3-digit. We observed that the evaluated segmentation-free approaches can achieves state-of-the-art performance without suffering the heavy burden of segmentation. In addition, the experiments confirm the importance of the Length classifier when recognizing strings of digits of unknown length. The information related to the number of digits in the string allow us to introduce some context to the problem, solving different confusions between isolated digits and touching components.

The remainder of this paper is organized as follows: Section \ref{sec:syntheticdata} presents the process used to create a synthetic dataset necessary to train our models on touching numeral strings with 2-, 3-, and 4-digits. Section \ref{sec:framework} describes the proposed segmentation free strategies, while Section \ref{sec:Experiments} presents the experiments performed to validate the proposed strategies for numeral string segmentation. Finally, the last section presents our conclusions and perspectives of future work. 

\section{Synthetic Data}
\label{sec:syntheticdata}

\begin{figure}[!htbp]
	\begin{center}
		\mbox{
			
			\subfigure[] {\epsfig {file=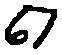, width=1.2cm}}
			\hspace{1cm}
			\subfigure[] {\epsfig {file=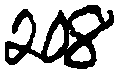, width=1.5cm}}
			\hspace{1cm}
			\subfigure[] {\epsfig {file=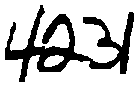, width=1.5cm}}
			
		}
		\caption{Synthetic data representing touching  numerical strings composed of (a) 2-digit, (b) 3-digit and (c) 4-digit.}
		\label{synthdataset:fig}
		\vspace{-5mm}
	\end{center}
\end{figure}

In order to efficiently learn representation from data, we had to rely on a considerable amount of samples. We thus created a synthetic dataset composed of touching numerical strings of sizes 2, 3, and 4. The strings are built by concatenating isolated digits of NIST SD19 \cite{NISTSD192016} through the algorithm described by Ribas et al. in \cite{Ribas2013}. Figure \ref{synthdataset:fig} shows some samples. The SD19 database, which is an update of SD3 and SD7, is provided by the American National Institute of Standards and Technology (NIST). This database contains the full page binary images of 3699 Handwriting Sample Forms (HSFs) and 814,255 segmented hand-printed digits and alphabetic characters from the forms.

To avoid building a biased dataset, we used the information on the authors available on the NIST SD19, such that digits from different authors were used exclusively for training, validation, and testing. Table \ref{tab:datadistrib} shows the purpose (training, validation, and testing), as well as the amount of data created\footnote{All the synthetic data is available upon request for research purposes at https://web.inf.ufpr.br/vri/databases-software/touching-digits/}. Isolated digits were extracted from NIST SD19. No data augmentation was necessary since more than 240,000 isolated digits are available in this dataset.

\begin{table}[!htb]
	\caption {Distribution of the data used for training and testing the classifiers. Samples are uniformly distributed among the classes.}
	\begin{center}
		\footnotesize
		\begin{tabular}{lrll} \hline
			\multicolumn{1}{c}{Length/Classes} &
			\multicolumn{1}{c}{Samples} &
			\multicolumn{1}{c}{Authors} &
			\multicolumn{1}{c}{Purpose} \\ \hline
			
			1 (Isolated digits) & 197,784 & 0000-2099 & Training \\  
			10 classes   	    & 23,384 & 3850-4099 & Validation  \\  
			& 23,621 & 3600-3849 & Testing \\  \hline
			2-Digit String      & 161,563 & 1000-1599 & Training \\  
			100 classes   		& 53,907 & 1600-1799 & Validation  \\  
			& 55,091 & 1800-1999 & Testing \\  \hline   
			3-Digit String      & 1,448,680 & 1000-1599 & Training \\  
			1000 classes   		& 484,346 & 1600-1799 & Validation  \\  
			& 491,749 & 1800-1999 & Testing \\  \hline   
			4-Digit String      & 100,000 & 1000-1599 & Training \\  
			*            		& 20,000 & 1600-1799 & Validation  \\  
			& 20,000 & 1800-1999 & Testing  \\ \hline 	
			
\multicolumn{4}{l}{\small *Data used to train the Length classifier. }	

			\label{tab:datadistrib}
		\end{tabular}
	\end{center}
	\vspace{-5mm}
\end{table}

\section{Segmentation-free Strategies}
\label{sec:framework}

\begin{figure}[htbp]
	\centering
	\epsfig{file=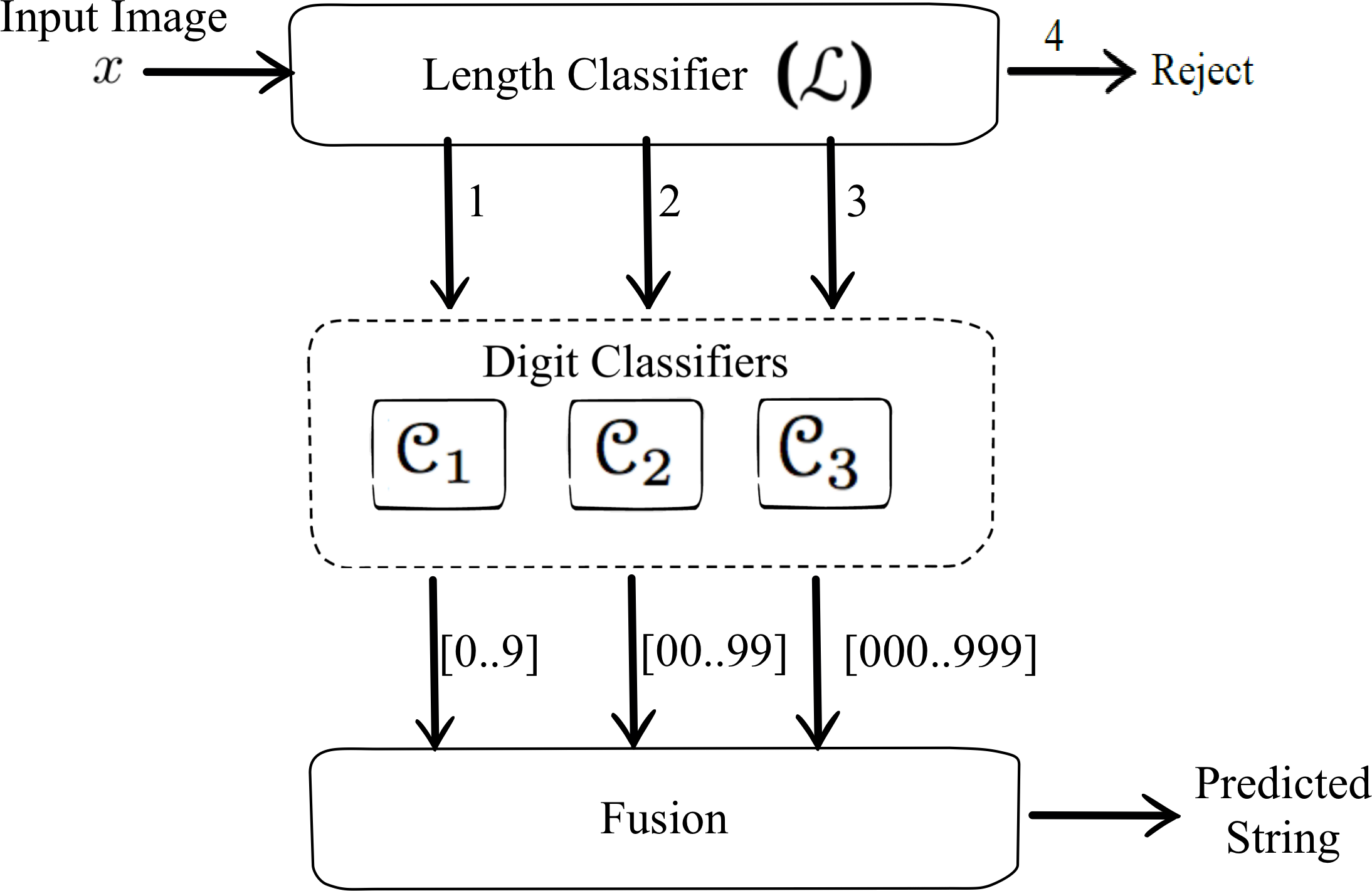, width=7cm}
	\caption{Segmentation-free framework proposed in \cite{Hochuli2018}.}
	\label{framework1:fig}
	\vspace{-3mm}
\end{figure}

The framework proposed in \cite{Hochuli2018} is depicted in Figure \ref{framework1:fig}. An image $x$ is first classified by the Length classifier ($\mathcal{L}$) which will assign to it a probability of having 1, 2, 3 or 4 touching digits. The digit classification module comprises three classifiers ($\mathcal{C}_{1}$, $\mathcal{C}_{2}$, $\mathcal{C}_{3}$) designed to discriminate 10 [$0 \ldots 9$], 100 [$00 \ldots 99$], and 1000 [$000 \ldots 999$] classes. The classifiers that will be used for a given image depends on the output of the Length Classifier. According to a fusion rule, more than one digit classifier may be invoked to mitigate any possible confusions.

The fusion rule used in this case considers the Top-2 outputs of $\mathcal{L}$. Let $\mathcal{L}^i(x) = p^i(x)$ be the probability of the input pattern $x$ be composed of $i, (i = 1,2,3,4)$ digits. Let $\mathcal{C}_{1}(x) = \max\limits_{0 \leq i \leq 9} p^i(x)$, $\mathcal{C}_{2}(x) = \max\limits_{0 \leq i \leq 99} p^i(x)$, and $\mathcal{C}_{3}(x) = \max\limits_{0 \leq i \leq 999} p^i(x)$ be the probability produced by 10-class, 100-class, and 1000-class classifiers, respectively, for the input pattern $x$. Let Top1($\mathcal{C}$) and Top2($\mathcal{C}$) be the functions that return the classes with first and second highest scores of a given classifier $\mathcal{C}$, respectively. Then, $x$ is assigned to the class $\omega \in [0...1110]$, according to Equation \ref{eq:prob1},

\begin{equation}
P(\omega|x)  \left \{ \begin{array}{ll}
\mbox{if } \mathcal{L}_(x) < T, & \max(\mathcal{C}_{Top1(\mathcal{L})}(x), \mathcal{C}_{Top2(\mathcal{L})}(x)) \\
\mbox{otherwise, }    & \mathcal{C}_{Top1(\mathcal{L})}(x)\\
\end{array}
\right.
\label{eq:prob1}
\end{equation}

\noindent where $T$ is a threshold defined empirically on the validation set. 

The justification for dealing with 1, 2, 3 touching digits is based on the fact that most of touching occurs between two digits and sometimes between three digits \cite{Wang00}. Strings composed of more than three touching digits are very rare in real problems and in the case of occurring $\mathcal{L}$ will reject them.

\begin{figure}[htbp]
	\centering
	\epsfig{file=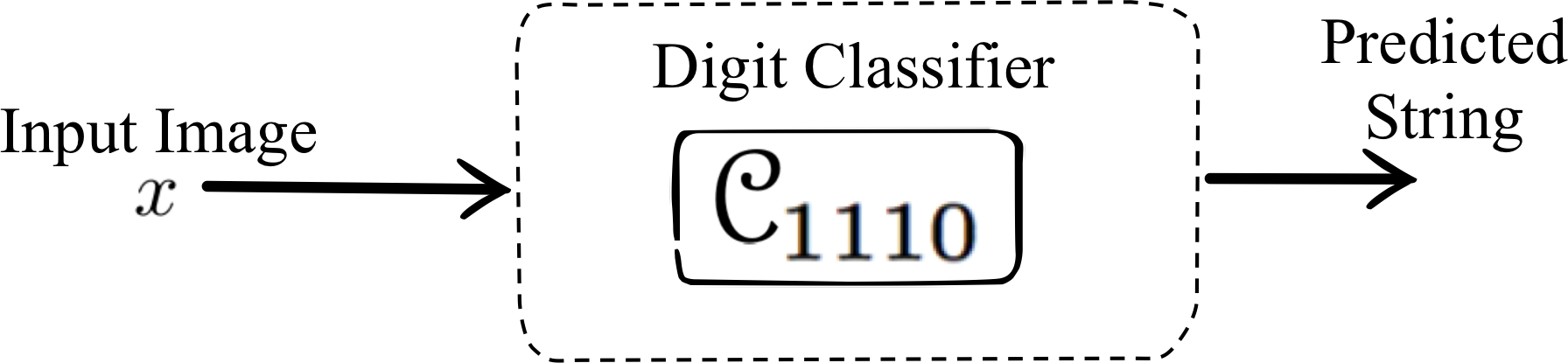, width=6cm}
	\caption{An end-to-end solution for touching digits.}
	\label{framework3:fig}
\end{figure}

This dynamic selection strategy has been proved quite efficient surpassing the results reported by all segmentation-based techniques reported in the literature. However, one may argue that an end-to-end solution with just one classifier ($\mathcal{C}_{1110}$) capable of recognizing those 1110 classes (10 isolated, 100 pairs, and 1000 triples), such as the one depicted in Figure \ref{framework3:fig}, is more elegant and easier to implement. In this case the classifier should encode not only the class of the object but also the length of the string.

\begin{figure}[htbp]
	\centering
	\epsfig{file=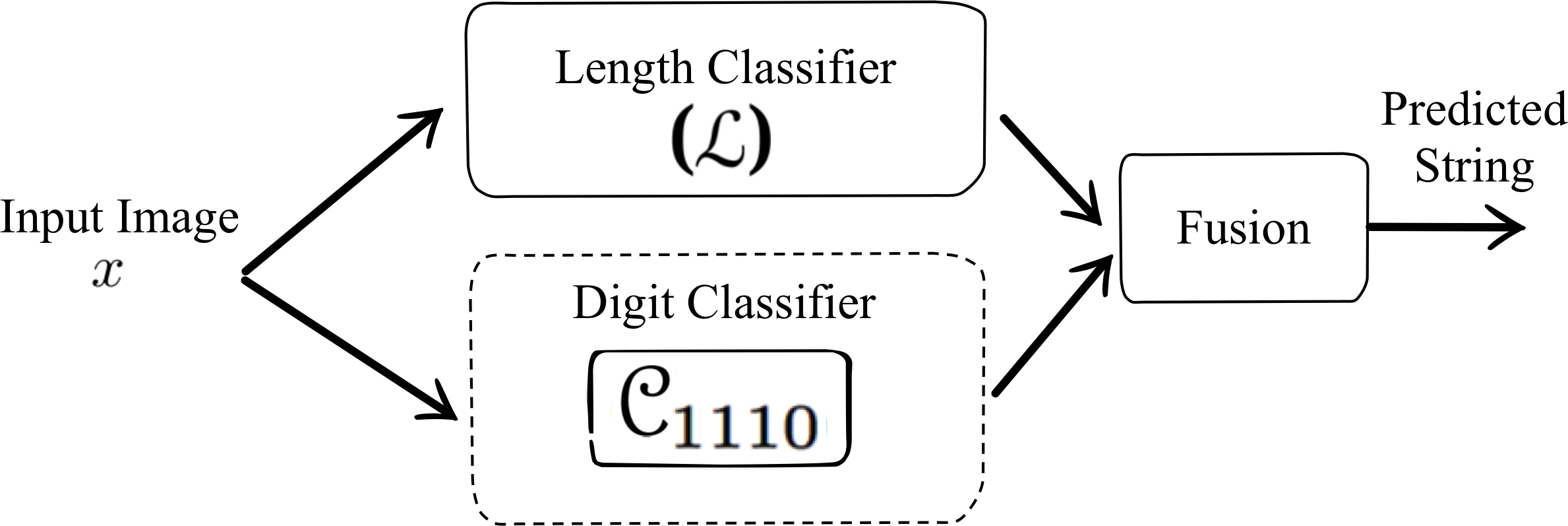, width=7.5cm}
	\caption{Single digit classifier combined with Length classifier $(\mathcal{L})$.}
	\label{framework2:fig}
	\vspace{-5mm}
\end{figure}

In fact the end-to-end solution is easier to implement, since it is based on a single classifier. However, as we will discuss in Section \ref{sec:Experiments}, this solution makes some confusions that could be easily solved having the information about the length of the string. With that in mind, we assess a third strategy (Figure \ref{framework2:fig}), in which we combine the output of the $\mathcal{C}_{1110}$ classifier with the Length classifier $(\mathcal{L})$. This approach uses the same fusion rule described earlier in this section. The difference is that the probability is produced by a single classifier instead of three.

Figure \ref{confusion_c1110:fig}a exemplifies the fusion process. In this case, $\mathcal{C}_{1110}$ misclassifies the input `60' by assigning it to class `610'. However, using the Top-1 output of $\mathcal{L}$, the correct class may be selected. In the case illustrated in Figure \ref{confusion_c1110:fig}b, because $\mathcal{L}$ produces a score smaller than $T$, the two outputs (Top-1 and Top-2) of $\mathcal{L}$ was used to solve the confusion. 

\subsection{Classifiers}
\label{classifiers:sec}
All the classifiers used in this work are CNNs that are constructed using multiple layers considering the following operations: convolutions, max-pooling, and dot products (fully-connected layers), where convolutional layers and fully connected layers have learnable parameters that are optimized during training. With the exception of the last layer in the network, after each learnable layer we apply ReLU non-linearity. The last layer uses the softmax non-linearity.

Training is performed with the Stochastic Gradient Descent (SGD) using back-propagation with mini-batches of 256 instances, a momentum factor of 0.9 and a weight decay of $5 \times 10^{-4}$. The learning rate is set to $10^{-2}$ in the beginning to allow the weights to quickly fit the long ravines in the weight space, after which it is reduced over the time (until  $5 \times 10^{-4}$) to make the weights fit the sharp curvatures. The network makes use of the well known cross-entropy loss function. 

In the present work, regularization was implemented through early-stopping, which prevents overfitting from interrupting the training procedure once the performance of the network on a validation set deteriorates. During training, the performance of the network on the training set will continue to improve, but its performance on the validation set will only improve up to a certain point, where the network starts to overfit the training data; at that point, the learning algorithm is terminated. To implement the CNN models we have used the Caffe framework \cite{jia2014caffe} on a NVidia GeForce GTX Titan Black GPU and NVidia GeForce GTX Titan Xp GPU\footnote{All trained classifiers are available for research purposes at https://web.inf.ufpr.br/vri/databases-software/touching-digits/}. 

\subsubsection{Length Classifier}
\label{CL:sec}

\begin{figure*}[htbp]
	\centering
	\includegraphics[width=.85\textwidth]{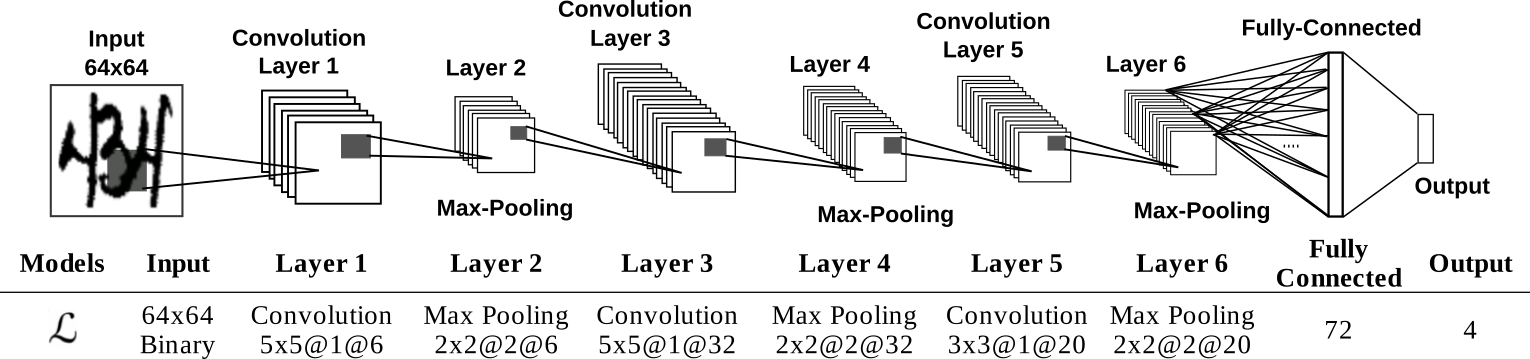}
	\caption{CNN architecture for $\mathcal{L}$. Layer parameters are represented as Kernel Size @ Stride @ Feature Maps.}
	\label{CNNLength:fig}
	\vspace{-3mm}
\end{figure*}

The length classifier $(\mathcal{L})$ was designed to predict the length of $x$. We have tested several different architectures for this classifier but the one that yielded the best results was based on the well-known LeNet 5 \cite{LeCun98}. The final architecture contained three convolutional layers followed by max pooling layers. This architecture, which was defined empirically on the validation set, is depicted in Figure \ref{CNNLength:fig}.

The classifier was trained using the protocol described in Section \ref{classifiers:sec} using 400,000, 79,157 and 79,742 samples (uniformly distributed) for training, validation, and testing, respectively. Using the Caffe framework and the hardware mentioned in Section \ref{classifiers:sec}, it took about 90 minutes to train this model over 30,000 iterations. Classifying a single input image takes about 0.4 milliseconds (ms). In our experiments, the best results were achieved when the input image was resized to $64 \times 64$ pixels. The recognition rate on the testing set was 98.4\% and 99.9\% for Top-1 and Top-2, respectively. Table \ref{tab:cm-cl} shows the confusion matrix.

\begin{table}[!h]
	\caption {Confusion matrix (\%) for the $\mathcal{L}$ on the testing set.}
	\begin{center}
		\begin{tabular}{ccccc} \hline
			\multicolumn{1}{c}{} &
			\multicolumn{1}{c}{(1)} &
			\multicolumn{1}{c}{(2)} &
			\multicolumn{1}{c}{(3)} &
			\multicolumn{1}{c}{(4)}  \\ \hline
			
			(1)  & 99.9 & 0.01 &       &     \\  
			(2)  & 0.02 & 99.2 &  0.07 &     \\  
			(3)  &      & 0.9  & 96.9  &  2.3\\  
			(4)  &      &      & 2.3   & 97.7 \\  \hline
			
			\label{tab:cm-cl}
			\vspace{-5mm}
		\end{tabular}
	\end{center}
\end{table}

\begin{figure}[htbp]
	\begin{center}
		\mbox{
			
			\subfigure[] {\epsfig {file=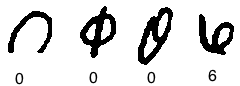, width=3cm}}
			\subfigure[] {\epsfig {file=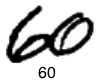, width=1.2cm}}
			\subfigure[] {\epsfig {file=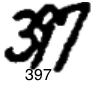, width=1.0cm}}
			\subfigure[] {\epsfig {file=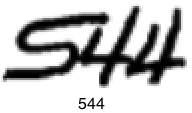, width=1.7cm}}
			
		}
		\caption{Some images misclassified by $\mathcal{L}$: (a) single digit classified as 2-digit string, (b) 2-digit classified as 3-digit string, (c) 3-digit classified as 2-digit string,  and (d) 3-digit classified as 4-digit string.}
		\label{erroLC:fig}
	\end{center}
\end{figure}

Analyzing the confusions resulting from $\mathcal{L}$ we conclude that the number and location of the vertical strokes seem to bear important information needed to determine the size of the string. For example, single digits that are classified as 2-digit string are often slashed zeros, zeros with missing parts, and the digit ``6'' similar to those presented in Figure \ref{erroLC:fig}a and b. Digits that are almost overlapping such as the ``3'' and ``9'' in Figure \ref{erroLC:fig}c and strings with several vertical strokes close together such as in the ``44'' in Figure \ref{erroLC:fig}d are also sources of confusion.

\subsubsection{Digit Classifiers}
\label{digitclassifiers:sec}

\begin{figure*}[htbp]
	\centering
		\includegraphics[width=.75\textwidth]{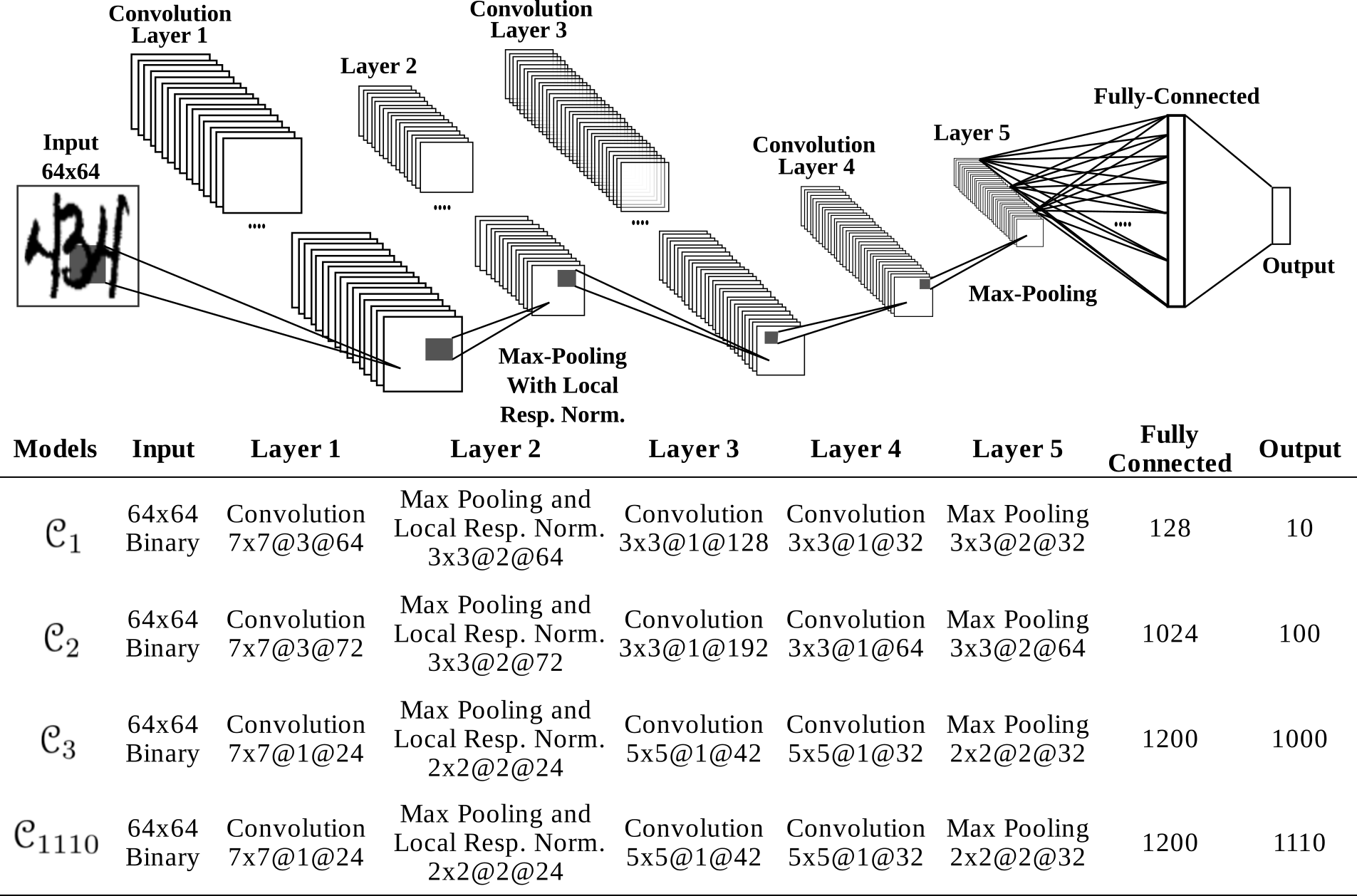}
	\caption{CNN architecture for digit classifiers. Layer parameters are represented as Kernel Size @ Stride @ Number of Feature Maps.}
	\label{CNNStrings:fig}
\end{figure*}

The classifiers $\mathcal{C}_{1}$, $\mathcal{C}_{2}$, $\mathcal{C}_{3}$, and $\mathcal{C}_{1110}$  presented in the previous section are based on the architecture depicted in Figure \ref{CNNStrings:fig}. The four CNNs, which also are based on the LeNet 5 \cite{LeCun98}, share the same structure but with different numbers of filters, kernel sizes, and strides. Figure \ref{CNNStrings:fig} summarizes the parameters used in all four classifiers, which were defined empirically on the validation set.

Table \ref{tab:digit-classifiers} shows the amount of data used for training, validation, and testing for all four classifiers. It also shows training (30,000 iterations) and classification time using the Caffe framework and the hardware mentioned in Section \ref{classifiers:sec}. 

\begin{table*}[!h]
	\caption {Data used to train the digit classifiers.}
	\begin{center}
		\begin{tabular}{cccccccc} \hline
			\multicolumn{1}{c}{Classifier} &
			\multicolumn{1}{c}{Number of} &
			\multicolumn{3}{c}{Amount of data ($\times 1000$) for} &
			\multicolumn{1}{c}{Source} &
			\multicolumn{1}{c}{Training} &
			\multicolumn{1}{c}{Classification} \\
			
			\multicolumn{1}{c}{} &
			\multicolumn{1}{c}{Classes} &
			\multicolumn{1}{c}{Train} &
			\multicolumn{1}{c}{Validation} &
			\multicolumn{1}{c}{Testing} &
			\multicolumn{1}{c}{} &
			\multicolumn{1}{c}{Time (min)} &
			\multicolumn{1}{c}{Time (ms)}  \\ \hline
			
			$\mathcal{C}_{1}$ & 10   & 197  & 23 & 23 & NIST SD 19      & 70$^1$ & 0.57$^1$\\  
$\mathcal{C}_{2}$ & 100  & 161  & 53 & 55 & Synthetic data  & 90$^1$ & 0.60$^1$\\  
$\mathcal{C}_{3}$ & 1000 & 1448 & 484 & 491 & Synthetic data& 200$^1$& 0.63$^1$\\  
			$\mathcal{C}_{1110}$ & 1110 & 1808 & 561 & 570 & Synthetic data& 183$^2$& 1.10$^1$\\  \hline
\multicolumn{4}{l}{\small (1) NVIDIA Titan Black GPU and (2) NVIDIA Titan Xp GPU}
            
			\label{tab:digit-classifiers}
				\vspace{-5mm}
		\end{tabular}
	\end{center}
\end{table*}

All four classifiers were trained using the protocol described in Section \ref{classifiers:sec} and yielded the accuracies reported in Table \ref{tab:performance-digit-classifiers}.

\begin{table}[!h]
\caption {Recognition rate of the Digit Classifiers on testing set.}
\begin{center}
\begin{tabular}{lcc} \hline
    \multicolumn{1}{c}{Classifier} &
    \multicolumn{1}{c}{Top 1} &
    \multicolumn{1}{c}{Top 2} \\ \hline
$\mathcal{C}_{1}$ & 99.6 & 99.9 \\  
$\mathcal{C}_{2}$ & 99.7 & 100.0 \\ 
$\mathcal{C}_{3}$ & 97.7 & 98.9 \\ 
$\mathcal{C}_{1110}$ & 95.9 & 98.6 \\ \hline
\label{tab:performance-digit-classifiers}
\end{tabular}
\end{center}
\vspace{-10mm}
\end{table}

\section{Experiments}
\label{sec:Experiments}

In order to validate the segmentation-free strategies we have used the 79,464 images of touching digits available in the Touching Pairs (TP) database \cite{Ribas2013}. This dataset allows us to better compare with the literature. We also perform experiment on the dataset described in Section \ref{sec:syntheticdata}, which contains single digits, 2-, and 3-digit connected. 

It is important to mention that, because these datasets contains only a single connected component per image, the pre-processing module was suppressed in those approaches. Also, the threshold value $T$ from Equation \ref{eq:prob1} was set to 0.95 according to the authors \cite{Hochuli2018}.
\subsection{TP dataset}
When evaluating the segmentation algorithms, the authors in \cite{Ribas2013} were interested in knowing whether or not the segmentation cuts produced by the algorithms were the good ones, independently of their quantity. For the algorithms based on the segmentation-recognition approach, this task is straightforward, since there is only one hypothesis to be assessed. For those algorithms based on over-segmentation, all the cuts must be assessed. In the latter case, the strategy used is as follows: if there are two digits among the hypotheses (using a classification engine) that match to the ground truth, the segmentation is considered successful. It is clear that this strategy considers the best case scenario since all misclassifications due to over- and under-segmentation are not considered.

\begin{figure}[htbp]
	\centering
	\epsfig{file=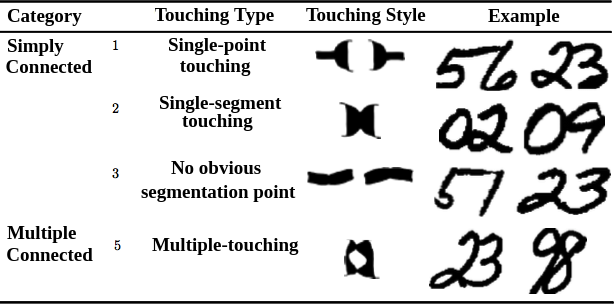, width=8cm}
	\caption{Types of connected numeral string (extracted from \cite{Ribas2013}).}
	\label{touching:fig}
	\vspace{-5mm}
\end{figure}

Table \ref{summarypairs:tab} summarizes the results reported in \cite{Ribas2013} and \cite{Gattal2015} where the authors compare several segmentation algorithms in terms of correct segmentation on the TP database. Besides the overall performance, this table also shows the performance depending on the connection types depicted in Figure \ref{touching:fig}.

\begin{table*} [htbp]
	\caption {Performance of the segmentation algorithms (reported in \cite{Ribas2013} and \cite{Gattal2015}), in terms of correct segmentation, on the TP Database.}
	\begin{center}
		\begin{tabular}{lccccccc} \hline
			\multicolumn{1}{c}{Method} &
			\multicolumn{1}{c}{Performance} &
			\multicolumn{4}{c}{Connection Type (\%)} &
			\multicolumn{1}{c}{Segmentation}\\ \cline{3-6}
			\multicolumn{1}{c}{} &
			\multicolumn{1}{c}{\%} &
			\multicolumn{1}{c}{I} &
			\multicolumn{1}{c}{II} &
			\multicolumn{1}{c}{III} &
			\multicolumn{1}{c}{V} &
			\multicolumn{1}{c}{Cuts}  \\ \hline
			
			Shi and Govindaraju (1997)                   & 59.30 & 68.31 & 59.72 & 60.35 & 25.44 & 1    \\
			Congedo et al. (1995)				 		 & 63.07 & 62.88 & 67.51 & 59.40 & 40.45 & 1   \\
			Lacerda and Mello (2013)   				     & 65.79 & 71.75 & 71.21 & 63.64 & 56.57 & 1    \\
			Elnagar and Alhajajj (2003)                  & 67.34 & 63.88 & 71.51 & 56.40 & 58.73 & 1    \\ 
			Pal et al. (2003)                            & 71.21 & 73.96 & 74.69 & 80.09 & 41.52 & 1    \\
			Oliveira et al. (2000)	             		 & 88.03 & 90.40 & 90.78 & 89.01 & 64.88 & 1    \\
			Fusijawa et al. (1992)                       & 89.85 & 95.45 & 91.27 & 83.57 & 63.72 & 3.66 \\
			Fenrich and  Krishnamoorthy (1990)           & 92.37 & 97.54 & 93.79 & 99.45 & 65.57 & 4.07 \\
			Gattal and Chibani (2015)                    & 93.24 & 96.67 & 93.75 & 99.68 & 77.58 & 24.11 \\
			Chen and Wang  (2000)                        & 93.80 & 97.87 & 94.23 & 97.55 & 76.76 & 45.40 \\ \hline
			
			Proposed End-to-end    								 & 94.37 & 94.33 & 95.13 & 96.13 & 91.90 & 0     \\ 
			Proposed End-to-end+$\mathcal{L}$					 & 96.05 & 95.95 & 96.87 & 98.03 & 93.35 & 0     \\ 
			Hochuli et al. (2018) 											 & 97.12 & 97.02 & 97.89 & 98.97 & 93.03 & 0     \\ \hline
			
		\end{tabular}
		\label{summarypairs:tab}
		\vspace{-5mm}
	\end{center}
\end{table*}

Table \ref{summarypairs:tab} also allows us to draw some conclusions. Algorithms based on a single segmentation hypothesis (segmentation cuts = 1) usually fail in more complex touching cases (e.g., type V) since a single segmentation cut is very often not enough to correctly split the digits. Algorithms based on multiple cuts, on the other hand, achieve better performance in terms of finding the correct segmentation cut, but with the computational cost of having to evaluate several hypotheses. 

In this context, the segmentation-free approaches compare favorably to traditional segmentation algorithms. In the End-to-End approach the expensive process of finding the segmentation cuts, filtering out unlikely hypotheses, and classifying the remaining ones is replaced by one classifier call ($\mathcal{C}_{1110}$). As we can see in Table \ref{summarypairs:tab}, this simple approach achieves 94.37\% of correct classification, which compares to the best methods reported in the literature, Chen and Wang \cite{Chen00} and Gattal et al.\cite{Gattal2015}. However, these two methods generate a large number of hypotheses, which makes them unfeasible for real applications due to the high computational cost.

\begin{figure*}[!htbp]
	\begin{center}
		
			\subfigure[]{\scalebox{0.45}{\epsffile{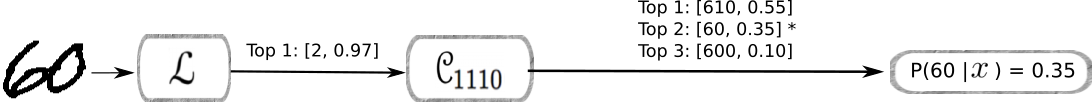}}} 
            \\
         	\subfigure[]{\scalebox{0.45}{\epsffile{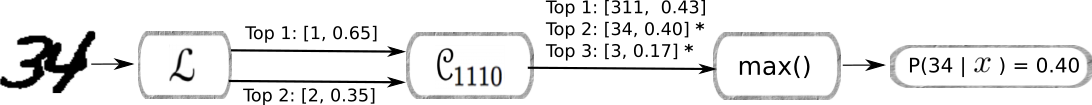}}}

		\caption{Confusions made by $\mathcal{C}_{1110}$ classifier - [Prediction, Probability]: Missed predictions for (a) `60' and (b) `34' were solved using information provided by the Length classifier ($\mathcal{L}$) and the fusion strategy from Eq \ref{eq:prob1}.}
		\label{confusion_c1110:fig}
	\end{center}
	\vspace{-8mm}
\end{figure*}

Figure \ref{confusion_c1110:fig} shows some confusions made by the $\mathcal{C}_{1110}$ classifier. Most of the errors are related to touching pairs confused with single digits or 3-digit strings. In light of this, the End-to-End approach could benefit somehow from the information provided by the Length classifier ($\mathcal{L}$), which is the strategy depicted in Figure \ref{framework2:fig}.

Using the End-to-End+($\mathcal{L}$), some of these confusions are solved increasing the recognition rate in about two percentage points (96.05\%). The total error (3.95\%) is caused in parts by $\mathcal{L}$ (1.76\%) and $\mathcal{C}_{1110}$ (2.79\%). In terms of computational cost, there is a little penalty since we have to add another classifier call and the fusion rule. However, compared to the traditional segmentation algorithms the cost is still negligible. 

\begin{figure}[htbp]
	\begin{center}

			\subfigure[]{\scalebox{0.55}{\epsffile{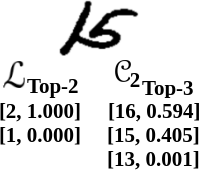}}} \quad
			\hspace{3mm}
			\subfigure[]{\scalebox{0.55}{\epsffile{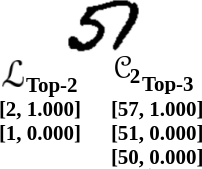}}} \quad 
		\caption{Confusions made by $\mathcal{C}_{2}$ classifier - [Prediction, Probability]: (a)`15' predicted as `16', (b) `51' predicted as `57'.}
		\label{confusionpairs:fig}
	\end{center}
\vspace{-6mm}
\end{figure}

Finally, the dynamic selection strategy presented in Figure \ref{framework1:fig} solves some of the confusions caused by the $\mathcal{C}_{1110}$. Instead of using a general-purpose classifier for 1-, 2-, and 3-digit strings, it divides the classification task into three parts, creating this way task-specific classifiers. In this experiment, though, only one of them is used along with $\mathcal{L}$. Assume that the size of the string is unknown, $\mathcal{C}_{2}$ is only used to classify the images that were assigned as 2-digit string by $\mathcal{L}$. This strategy reaches the highest performance (97.12\%). Comparing to the End-to-End+($\mathcal{L}$), the classification error was reduced from 2.79\% to 1.10\%. Figure \ref{confusionpairs:fig} shows some images that were misclassified by $\mathcal{C}_{2}$. As reported in Table \ref{summarypairs:tab}, the poorest performance (93\%) is achieved on type V (multiple touching), which shows the highest variability. However, when compared to others, such a performance is outstanding.

\subsection{Synthetic Data}

In the previous experiments only touching pairs were considered so that we could compare the segmentation-free approaches with the literature. Besides, segmenting touching pairs is the main bottleneck of any string recognition system. However, in several cases, a digit string is composed mainly by isolated digits and sometimes it may contain three or more digits connected. In this section we assess the segmentation-free approaches on the synthetic data described in Section \ref{sec:syntheticdata}, which contains over 570,000 images of isolated digits, touching strings of 2- and 3-digits. 

\begin{figure}[htbp]
	\centering
	\epsfig{file=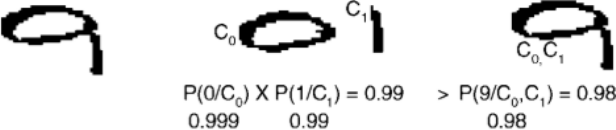, width=6.0 cm}
	\caption{Misclassification caused by over-segmentation (extracted from \cite{Oliveira02b}).}
	\label{overseg:fig}
\end{figure}

One may argue that recognition of isolated digits is a problem already solved since the literature shows accuracy close to 100\% \cite{Ciresan2012,Sabour2017}. It is worth remembering, however, that the lack of context in digit string recognition makes the problem more challenging since an image may contain an isolated digit or several digits connected. To deal with this problem, heuristic-based segmentation algorithms rely on over-segmentation to maximize the chances of finding the correct segmentation point, even when segmentation is not necessary. This strategy has a downside, i.e., isolated digits that do not need segmentation may be segmented and the over-segmented pieces recognized with high probabilities. This is exemplified in Figure \ref{overseg:fig} where the digit ``9'' was over-segmented into two parts, which were recognized as ``0''and ``1'' with high probability.

Table \ref{summarysynthetic:tab} shows the results of the three segmentation-free approaches discussed in this work.

\begin{table} [htbp]
	\caption {Performance of the segmentation-free approaches on the synthetic data.}
	\begin{center}
		\begin{tabular}{lccc} \hline
			\multicolumn{1}{c}{Method} &
			\multicolumn{1}{c}{Single digit} &
			\multicolumn{1}{c}{2-digit} &
			\multicolumn{1}{c}{3-digit} \\ \hline
			
			End-to-end    								 & 97.68 & 94.09 & 96.05  \\ 
			End-to-end+$\mathcal{L}$					 & 98.73 & 96.82 & 95.50      \\ 
			Hochuli et al. (2018)										 & 99.56 & 99.00 & 94.88   \\ \hline
			
		\end{tabular}
		\label{summarysynthetic:tab}
	\end{center}
\end{table}

The results achieved by both End-to-end+$\mathcal{L}$ and Hochuli et. al. \cite{Hochuli2018} corroborate to the importance of the Length classifier when recognizing strings of digits of unknown length. Several confusions between isolated digits and touching digits are solved by using the information about the number of digits in the string. 

\begin{figure}[!htbp]
	\begin{center}
		\mbox{
        \subfigure[]{\scalebox{0.55}{\epsffile{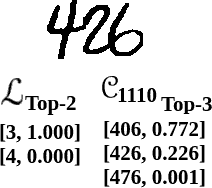}}} \quad
        \hspace{4mm}
		\subfigure[]{\scalebox{0.55}{\epsffile{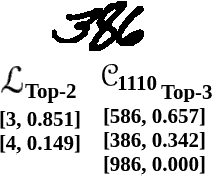}}} 
		 					
		}
		\caption{Confusion made by $\mathcal{C}_{1110}$ - [Predicion,Probability]: (a) `426' predicted as `406' and (b) `386' predicted as  `586'.}
		\label{confusion_c1110_3dig:fig}
	\end{center}
\end{figure}

In the case of 3-digit strings, which are not very often in real datasets, most of the confusions occur intra-class, e.g., ``426'' confused with ``406'' depicted in Figure \ref{confusion_c1110_3dig:fig}. Since strings with three touching digits contain more information to encode the size of the string the Length classifier does not contribute to improve the recognition rate. On the other hand, segmentation-based approaches will suffer with a higher number of hypotheses to be assessed. 

\section{Conclusion}
\label{conclusion:sec}

Since segmentation of touching digits remains a challenge for handwritten numeral recognition, in this paper we have presented segmentation-free approaches corroborating with the recent work \cite{Hochuli2018} that achieved state-of-art performance through a dynamic selection strategy based on four deep learning models. Towards an end-to-end solution, we have implemented a touching digit classifier that is capable to discriminate 1110 classes (10 for isolated, 100 for pairs and 1000 for triples). Using a strong experimental protocol, the proposed approaches surpass segmentation-based methods bringing up a new perspective to the problem. Further analysis confirmed that introducing context information related to the length of string predicted by a trained classifier is an useful strategy to solving some confusions made by digit classifiers. For future works, we are developing an approach that combine length and digits classifiers into an end-to-end solution.

\section*{Acknowledgements}

This research has been supported by The National Council for Scientific and Technological
Development (CNPq) grant  303513/2014-4. In addition, we gratefully acknowledge the support of NVIDIA Corporation with the donation of the Titan Xp GPU used for this research.

\bibliographystyle{IEEEtran}
\bibliography{refer1}

\begin{thebibliography}{10}
\providecommand{\url}[1]{#1}
\csname url@samestyle\endcsname
\providecommand{\newblock}{\relax}
\providecommand{\bibinfo}[2]{#2}
\providecommand{\BIBentrySTDinterwordspacing}{\spaceskip=0pt\relax}
\providecommand{\BIBentryALTinterwordstretchfactor}{4}
\providecommand{\BIBentryALTinterwordspacing}{\spaceskip=\fontdimen2\font plus
\BIBentryALTinterwordstretchfactor\fontdimen3\font minus
  \fontdimen4\font\relax}
\providecommand{\BIBforeignlanguage}[2]{{%
\expandafter\ifx\csname l@#1\endcsname\relax
\typeout{** WARNING: IEEEtran.bst: No hyphenation pattern has been}%
\typeout{** loaded for the language `#1'. Using the pattern for}%
\typeout{** the default language instead.}%
\else
\language=\csname l@#1\endcsname
\fi
#2}}
\providecommand{\BIBdecl}{\relax}
\BIBdecl

\bibitem{Ribas2013}
F.~C. Ribas, L.~S. Oliveira, A.~S. Britto, and R.~Sabourin, ``Handwritten digit
  segmentation: A comparative study,'' \emph{International Journal on Document
  Analysis and Recognition}, vol.~16, no.~2, pp. 567--578, 2013.

\bibitem{Vellasques2008}
E.~Vellasques, L.~S. Oliveira, A.~S. Britto, A.~Koerich, and R.~Sabourin,
  ``Filtering segmentation cuts for digit string recognition,'' \emph{Pattern
  Recognition}, vol.~41, no.~10, pp. 3044--3053, 2008.

\bibitem{Matan92}
O.~Matan, J.~C. Burges, Y.~LeCun, and J.~S. Denker, ``Multi-digit recognition
  using a space displacement neural network,'' in \emph{Advances in Neural
  Information Processing Systems}, J.~E. Moody, S.~J. Hanson, and R.~L.
  Lippmann, Eds.\hskip 1em plus 0.5em minus 0.4em\relax Morgan Kaufmann, 1992,
  vol.~4, pp. 488--495.

\bibitem{LeCun98}
Y.~LeCun, L.~Bottou, Y.~Bengio, and P.~Haffner, ``Gradient-based learning
  applied to document recognition,'' \emph{Procs of IEEE}, vol.~86, no.~11, pp.
  2278--2324, 1998.

\bibitem{Choi99}
S.~Choi and I.~Oh, ``A segmentation-free recognition of two touching numerals
  using neural networks,'' in \emph{Proc. of 5$^{th}$ International Conference
  on Document Analysis and Recognition}, Bangalore, India, 1999, pp. 253--256.

\bibitem{Ciresan2008}
D.~Ciresan, ``Avoiding segmentation in multi-digit numeral string recognition
  by combining single and two-digit classifiers trained without negative
  examples,'' in \emph{10th International Symposium on Symbolic and Numeric
  Algorithms for Scientific Computing}, 2008, pp. 225--230.

\bibitem{Hochuli2018}
A.~G. Hochuli, L.~S. Oliveira, A.~S. Britto, and R.~Sabourin, ``Handwritten
  digit segmentation: Is it still necessary?'' \emph{Pattern Recognition},
  vol.~78, pp. 1 -- 11, 2018.

\bibitem{BRITTO2014}
A.~S. Britto, R.~Sabourin, and L.~S. Oliveira, ``Dynamic selection of
  classifiers—a comprehensive review,'' \emph{Pattern Recognition}, vol.~47,
  no.~11, pp. 3665 -- 3680, 2014.

\bibitem{CRUZ2018}
R.~M. Cruz, R.~Sabourin, and G.~D. Cavalcanti, ``Dynamic classifier selection:
  Recent advances and perspectives,'' \emph{Information Fusion}, vol.~41, pp.
  195 -- 216, 2018.

\bibitem{NISTSD192016}
P.~J. Grother, \emph{NIST Special Database 19 - Handprinted forms and
  characters database}, NIST, 2016.

\bibitem{Wang00}
X.~Wang, V.~Govindaraju, and S.~N. Srihari, ``Holistic recognition of
  handwritten character pairs,'' \emph{Pattern Recognition}, vol.~33, no.~12,
  pp. 1967--1973, 2000.

\bibitem{jia2014caffe}
Y.~Jia, E.~Shelhamer, J.~Donahue, S.~Karayev, J.~Long, R.~Girshick,
  S.~Guadarrama, and T.~Darrell, ``Caffe: Convolutional architecture for fast
  feature embedding,'' \emph{arXiv preprint arXiv:1408.5093}, 2014.

\bibitem{Gattal2015}
A.~Gattal and Y.~Chibani, ``{SVM}-based segmentation-verification of
  handwritten connected digits using the oriented sliding window,''
  \emph{International Journal of Computational Intelligence and Applications},
  vol.~14, no.~1, pp. 1--17, 2015.

\bibitem{Chen00}
Y.~K. Chen and J.~F. Wang, ``Segmentation of single- or multiple-touching
  handwritten numeral string using background and foreground analysis,''
  \emph{IEEE Trans. on Pattern Analysis and Machine Intelligence}, vol.~22,
  no.~11, pp. 1304--1317, 2000.

\bibitem{Ciresan2012}
D.~Ciresan, U.~Meier, and J.~Schmidhuber, ``Multi-column deep neural networks
  for image classification,'' in \emph{2012 IEEE Conference on Computer Vision
  and Pattern Recognition}, June 2012, pp. 3642--3649.

\bibitem{Sabour2017}
S.~Sabour, N.~Frosst, and G.~Hinton, ``Dynamic routing between capsules,'' in
  \emph{Advances in Neural Information Processing Systems 30 (NIPS 2017)},
  2017.

\bibitem{Oliveira02b}
L.~S. Oliveira, R.~Sabourin, F.~Bortolozzi, and C.~Y. Suen, ``Automatic
  recognition of handwritten numerical strings: A recognition and verification
  strategy,'' \emph{IEEE Trans. on Pattern Analysis on Machine Intelligence},
  vol.~24, no.~11, pp. 1438--1454, 2002.

\end{thebibliography}
\end{document}